%% file: 00_main.tex
\documentclass{article}

\usepackage[preprint]{neurips_2025}

\input{05_declarations}

\title{Streaming Sequence-to-Sequence Learning with Delayed Streams Modeling}

\author{%
  Neil Zeghidour$^{*}$ $\quad$
  Eugene Kharitonov $\quad$
  Manu Orsini \\ 
 \textbf{ V\'aclav Volhejn} $\quad$
  \textbf{Gabriel de Marmiesse} $\quad$
  \textbf{Edouard Grave} \\
  \textbf{Patrick P\'erez} $\quad$
  \textbf{Laurent Mazar\'e} $\quad$ \textbf{Alexandre D\'efossez}$^{*}$ \vspace{1mm} \\
  {$^*$Equal contribution}\\
  Kyutai \\
  \texttt{\{neil, eugene, alex\}@kyutai.org}
}

\begin{document}

\maketitle

\begin{abstract}
We introduce Delayed Streams Modeling (DSM), a flexible formulation for streaming, multimodal sequence-to-sequence learning. Sequence-to-sequence generation is often cast in an offline manner, where the model consumes the complete input sequence before generating the first output timestep. Alternatively, streaming sequence-to-sequence rely on learning a policy for choosing when to advance on the input stream, or write to the output stream. DSM instead models already time-aligned streams with a decoder-only language model. By moving the alignment to a pre-processing step,
and introducing appropriate delays betweem streams, DSM provides streaming inference of arbitrary output sequences, from any input combination, making it applicable to many sequence-to-sequence problems.
In particular, given a text and audio stream, automatic speech recognition (ASR) corresponds to the text stream
    being delayed, while the opposite gives a text-to-speech (TTS) model. We perform extensive experiments for these two major sequence-to-sequence tasks, showing that DSM provides state-of-the-art performance and latency while supporting arbitrary long sequences, being even competitive with offline baselines. Code, samples and demos are available at \href{https://github.com/kyutai-labs/delayed-streams-modeling/}{github.com/kyutai-labs/delayed-streams-modeling}.
    
\end{abstract}

\section{Introduction}
\label{sec:intro}
\looseness=-1
We are interested in streaming sequence-to-sequence (seq2seq) learning, i.e.\ predicting an output sequence as we process an input sequence synchronously, as opposed to offline seq2seq where inputs are recorded entirely before producing the output sequence. The latter class of offline models was introduced for a diverse set of tasks such as handwriting recognition~\citep{graves_ctc}, automatic speech recognition (ASR)~\citep{graves_ctc} or machine translation~\citep{bahdanau_attention,sutskever2014sequence}, by designing modality-dependent input encoders, typically coupled with a text decoder~\citep{lstm}. Although this asymmetry between input processing and output generation facilitated the adoption of this framework in many tasks, it also led to a divergence of model architectures across modalities. As an example, a Tacotron text-to-speech (TTS) model~\citep{tacotron} would differ from an ASR model such as LAS~\citep{chan2016listen}. %
The advent of decoder-only Transformers~\citep{attentionvaswani,radford2018improving} for text language modeling reduced the gap between input and output processing by allowing a single model to process a simple concatenation of tokens. In parallel, neural compression algorithms that can transform images~\citep{vqvae,Esser2020TamingTF} and audio~\citep{soundstream,encodec} into discrete tokens analogous to text allowed integrating these modalities along text sequences. Thus, a decoder-only model can be used for seq2seq tasks such as ASR~\citep{audiopalm}, TTS~\citep{valle,speartts}, spoken dialogue~\citep{moshi}, visual understanding~\citep{paligemma} or image generation~\citep{dalle}. Furthermore, inputs and outputs are interchangeable in this framework, meaning a single model can be trained for generation in both directions: AudioPALM~\citep{audiopalm} performs TTS and ASR, while CM3Leon~\citep{Yu2023ScalingAM} provides both image captioning and generation. Yet, a major limitation of these decoder-only approaches is their incompatibility with streaming. First, their prefix-based formulation requires access to the full input sequence before generation, which prevents real-time inference and inherently limits the maximum input length. Second, modalities operate at differing framerates: audio or video tokens are typically sampled regularly, while text tokens represent linguistic units pronounced with varying durations. This prevents applications such as meeting transcription or continuous translation. 

Another popular approach is to learn an alignment policy between modalities, using architectures such as Transducers~\citep{graves2012sequence,zhang2020transformer}, or specific attention formulations~\citep{monotonic_attention,guo2024decoder}. 
At inference, the policy decision will change which modules to execute at each step, which is detrimental to batching. Besides, learning the policy requires train-time exploration, which can be costly. As noted by~\citet{ma-etal-2019-stacl}, a simple \emph{wait-k} policy can be used, especially for same modality sequence-to-sequence modeling.

\looseness=-1
In this work, we present Delayed Streams Modeling (DSM), a framework for streaming sequence-to-sequence learning across modalities. We make a simplifying assumption compared with previous \emph{wait-k} based methods~\citep{ma2021streaming,chen-etal-2021-direct}, namely that both modalities are aligned to a shared framerate as a pre-processing step.
DSM uses a decoder-only model to process as many parallel token streams as there are I/O sequences. This multistream architecture, introduced by~\citet{moshi}, allows for a synchronous autoregressive modeling of aligned sequences which---when coupled with a finite context---provides real-time, streaming generation over infinite input sequences. Moreover, by operating at a constant framerate, DSM allows for batching, a feature rarely provided by streaming models. The second key component of DSM, inspired by the \emph{wait-k} policy~\citep{ma-etal-2019-stacl}, is the introduction of a delay between streams to control the quality/latency trade-off: shifting a sequence B such that it is delayed w.r.t.\ sequence A allows for a better prediction of the former based on the latter. With appropriate delays, a DSM model can be trained to continuously predict any combination of output sequences from any combination of input sequences. 
To illustrate the abilities of the DSM framework, we train speech-text models for ASR and TTS. We show how DSM provides a state-of-the-art tradeoff between latency---as low as a few hundred milliseconds---and quality, while providing long-form synthesis and transcription, along with precise word timestamps that locate where they are pronounced.

\section{Related Work}
\label{sec:related}
\textbf{Streaming Sequence-to-Sequence Learning.}\ 
\looseness=-1
Most streaming seq2seq literature has focused on speech-to-text tasks, in particular ASR~\citep{streaming_asr} and translation~\citep{xue2023weakly,seamless}. Monotonic~\citep{monotonic_attention, mocha} and local~\citep{local_attention} attention respectively allow for causal attention of outputs with respect to inputs along with handling arbitrarily long sequences. A common limitation of streaming models is their incompatibility with batching when using an inference policy~\citep{seamless}, or the lack of symmetry meaning that specific models must be used for speech-to-text~\citep{streaming_asr} and text-to-speech~\citep{tacotron}.
Previous approaches using Transformer decoder-only models~\citep{guo2024decoder,chen2024streaming} typically require non-standard attention, and separate calls to the backbone per modality.
In contrast, DSM allows for batching and accelerated inference, using only standard attention, with all modalities fused to limit the number of steps in the backbone decoder.
In the context of this paper, this allows DSM to be trained for state-of-the-art ASR or TTS (see Figure~\ref{fig:concept}), as shown in Section~\ref{sec:experiments}, with its performance being even competitive with offline approaches.

\textbf{Multimodal language models.}\ 
\looseness=-1
Transformer-based autoregressive models are the current main approach to sequence-to-sequence problems. They were introduced by~\citet{attentionvaswani} for machine translation, and were soon extended to multimodal tasks, such as ASR~\citep{whisper} or visual understanding~\citep{flamingo}, by designing modality-specific encoders. More recently, neural codecs have provided compact, discrete representations of images~\citep{Esser2020TamingTF} and audio~\citep{soundstream} that remove the need for modality-specific encoders inside the generative model, while providing a symmetrical processing of inputs and outputs which allows performing bidirectional tasks (e.g.\ speech-to-text and text-to-speech~\citep{audiopalm}) with a single architecture. ~\citet{moshi} introduce a multistream decoder architecture for spoken dialogue, which predicts text and audio tokens in a streaming fashion, later applied by~\citet{hibiki} to real-time speech translation. In this work we extend the approach of~\citet{moshi}, in order to reach state-of-the-art performance on the two most competitive speech-text tasks, namely ASR and TTS.
Moreover, while~\citet{moshi} and~\citet{hibiki} operate with a delay specified before training, we propose delay conditioning for inference-time latency control without retraining.
Our TTS covers both monologue and controllable dialog generation, a topic that was studied by CoVoMix~\citep{zhang2024covomix}, although at a lower sample rate (8 kHz) and not streaming.

\input{30_method}
\section{Experiments}
\label{sec:experiments}

\begin{table}[t]
\centering
\caption{\textbf{Short-form ASR performance}. We report Word Error Rates (WER, \%) for \oursASR{} and selected non-streaming baselines from the OpenASR leaderboard, along with streaming baselines.}
\label{tab:openasr-leaderboard}
\begin{sc}
\resizebox{\textwidth}{!}{
\begin{tabular}{lc|cccccccc}
\toprule
\textbf{Model} & \textbf{Avg.} %
& \textbf{AMI} & \textbf{Earnings22} & \textbf{Gigaspeech} & \textbf{LS Clean} & \textbf{LS Other} & \textbf{SPGISpeech} & \textbf{TED-LIUM} & \textbf{Voxpopuli}
\\
\midrule
\multicolumn{10}{c}{\textit{Non-streaming}} \\
\midrule[0.3pt]
Whisper medium.en & 8.1 %
& 16.7 & 12.6 & 11.0 & 3.0 & 5.9 & 3.3 & 4.1 & 9.6 \\
Whisper Large-v3 & 7.5 %
& 16.0 & 11.3 & 10.0 & 2.0 & 3.9 & 2.9 & 3.9 & 9.5 \\
Voxtral Mini & 7.1 & 16.3 & 10.7 & 10.2 & 1.9 & 4.1 & 2.4 & 3.7 & 7.1 \\
ElevenLabs Scribe & 6.9 %
& 14.4 & 12.1 & 9.7 & 1.8 & 3.3 & 3.3 & 3.2 & 7.2 \\
CrisperWhisper & 6.7 %
& \textbf{8.7} & 12.9 & 10.2 & 1.8 & 4.0 & 2.7 & 3.2 & 9.8 \\
Canary-Flash & 6.4 %
& 13.1 & 12.8 & 9.9 & \textbf{1.5} & \textbf{2.9} & 2.0 & 3.1 & \textbf{5.6} \\
Phi-4 Multimodal & {6.1} %
& 11.5 & \textbf{10.5} & 9.8 & 1.7 & 3.8 & 3.1 & 2.9 & 5.9 \\
Parakeet-TDT-v2 & {6.1} %
&  11.2 & 11.2 & {9.7} & 1.7 & 3.2 & 2.2 & 3.4 & 6.0 \\
Canary-Qwen-2.5B & \textbf{5.6} &  10.2 &\textbf{10.5}& \textbf{9.4} & 1.6 & 3.1 & \textbf{1.9} & \textbf{2.7} & 5.7 \\

\midrule
\multicolumn{10}{c}{\textit{Streaming}} \\
\midrule[0.3pt]
Whisper medium.en & 9.0  & 22.1 & 13.4 & 10.4 & 3.0 & 6.2 & 3.7 & 4.7 & 8.6 \\
Whisper large-v3 & 9.4  & 18.4 & 11.0 & 10.0 & 8.4 & 12.6 & 3.2 & 3.8 & 7.9 \\
SeamlessStreaming & 19.7 & 45.0 & 31.8 & 21.6 & 6.8 & 10.6 & 15.4 & 12.4 & 13.9 \\
Parakeet-TDT-v2 &  7.0 & \textbf{11.9} & 11.8 & 10.2 & 3.3 & 4.7 & 3.4 & 3.9 & \textbf{6.6} \\ 
\midrule[0.01pt]
\oursASR & \textbf{6.4} & {12.2} & \textbf{11.0} & \textbf{9.8} & \textbf{1.7} & \textbf{4.3} & \textbf{2.0} & \textbf{3.4} & {6.8} \\
\bottomrule
\end{tabular}
}
\end{sc}
\end{table}

\subsection{Architectural hyperparameters}
\label{sec:hyperparams}
\looseness=-1
We use a Transformer~\citep{attentionvaswani} backbone with RoPE positional encoding~\citep{su2024roformer}. For the \oursTTS{} experiments, we use a 1B parameters backbone with a 2048 dimension latent space, GLU feed-forward units, 16 layers, and 16 heads of attention.
The TTS model also receives the speaker embedding through cross-attention layers.
The sampler is a Transformer along the codebook $Q$ dimension described in Section~\ref{sec:representations}, with no context over the time axis, with a dimension of 1024, 4 layers for each codebook, with a linear layer to estimate the logits. While \citet{moshi} used a different set of parameters for each codebook in this Transformer over $Q$, we follow \citet{hibiki}, using only individual set of parameters for the first 8 codebooks, then sharing parameters per group of 8 codebooks, for a total of 1.8B parameters with the backbone.
The text tokenizer is trained on bilingual French/English data, with a cardinality $N_t^\text{tts}=8000$.
The model uses a delay $\tau$ of 1.6s, or 16 steps, and a lookahead stream with $l = 2$ (Section~\ref{sec:dsm_tts}).

\looseness=-1
The \oursASR{} uses a 2.6B parameters backbone, with 2048 dimensions, 48 layers, and 32 attention heads, and a linear to predict the logits over the text vocabulary, with a cardinality $N_t^\text{asr} = 4000$, trained for English only. We experiment with two flavors of the ASR model: one uses a single fixed delay value for all examples, the other has a variable delay $\tau$ which is sampled per batch item in a range of $[0.25, 4]$s (Section~\ref{sec:dsm_asr}).

\subsection{Training protocol}
\label{sec:training_protocol}
\looseness=-1
\textbf{Pretraining.}\
We use an audio collection of 2.5 million hours of publicly available audio content in English and French transcribed with \texttt{whisper-timestamped}. Given the synthetic nature of text transcripts, this phase amounts to hard distillation of \texttt{whisper-timestamped}. We train \oursASR{} on random 90s segments for 1.6M steps, on 48 H100s. %
\oursTTS{} is trained on 150s audio extracts, on 32 H100s, 750k updates with batch size 64.
We use AdamW~\citep{adamw}, a cosine learning rate schedule with linear warmup, with an initial rate of $2\cdot\mathrm{10}^{-4}$ for the TTS, and 
 $4\cdot\mathrm{10}^{-4}$ for the ASR, and a weight decay of $0.1$.

\looseness=-1
\textbf{Finetuning (\oursASR{}).}\ We then finetune the model on a collection of public datasets with ground-truth transcripts, described in Appendix~\ref{sec:app:short-form} and totaling 28k hours. This training stage lasts for 100k updates with batches of 128 examples, using 16 H100s.  We augment training by using the codebook dropout~\citep{moshi}.
We then adapt the model to long-form inputs, which most public datasets lack, by constructing a long-form mixture described in Appendix~\ref{sec:app:long-form}. We run this stage for \oursASR{} for 25k updates with batch size 32, using 16 H100s. %

\textbf{Finetuning (\oursTTS{}).}\ 
 For safety reasons, we want this open-source model to be usable only with proper speaker embeddings from our speaker encoder, which we keep closed-source. This is not compatible with the CFG formula given by Eq.~\eqref{eq:cfg}, which requires the model to be usable unconditionally.
 Thus, we fine-tune the TTS with distillation of the CFG step, using an extra tensor conditioning summed with each backbone input to indicate value for $\alpha \in \{1, 1.5, 2, 2.5, 3, 3.5, 4\}$.
 The model is fined tuned for 2400 updates with a learning rate of $1\cdot 10 ^{-6}$.

\subsection{Automatic Speech Recognition}

\begin{table}[t]
\centering
\tiny
\caption{\textbf{Long-form ASR performance}. We report Word Error Rates (WER, \%) across four long-form datasets for \oursASR{} and a set of streaming and non-streaming baselines.}
\label{tab:longform-asr}
\begin{sc}
\begin{tabular}{lc|cccc}
\toprule
\textbf{Model} & \textbf{Avg.} & \textbf{TED-LIUM} & \textbf{Meanwhile} & \textbf{Rev16} & \textbf{Earnings21} \\
\midrule
\multicolumn{6}{c}{\textit{Non-streaming}} \\
\midrule[0.3pt]
distil-large v2 & 8.7 & 3.7 & 7.8 & 12.2 & 11.2 \\ %
whisper-large-v2 & 9.0 & 4.4 & 6.3 & 13.6 & 11.8 \\ %
\midrule
\multicolumn{6}{c}{\textit{Streaming}} \\
\midrule[0.3pt]
Whisper medium.en & 9.0 & 3.9 & 6.7 & 13.0 & 12.5 \\ %
Whisper large-v3 & 8.1 & 3.4 & 6.1 & {11.4} & 11.4 \\ %
Parakeet-0.6B-tdt-v2 & \textbf{7.8} & 3.7 & \textbf{5.0} &\textbf{11.0} & 11.3  \\
\midrule
\oursASR & {7.9} & \textbf{2.9} & {5.7} & 12.3 & \textbf{10.6} \\ %
\bottomrule
\end{tabular}
\end{sc}
\end{table}

\looseness=-1
We evaluate \oursASR{} (with a default fixed delay of 2.5s) in terms of transcription quality, latency, and timestamps precision. We consider short-form transcription (shorter than 30s), as it is the focus of the OpenASR Leaderboard~\citep{open-asr-leaderboard}. We also look at streaming inference for long-form transcription (up to 2 hours).

\textbf{Baselines.}\ 
We benchmark \oursASR{} against leading models of the OpenASR Leaderboard, including Whisper~\citep{whisper}, 
Canary-Flash~\citep{canary}, Phi-4 Multimodal Instruct~\citep{phi4}, Parakeet~\citep{parakeet}, Voxtral Mini~\citep{voxtral} and as well as the closed-source ElevenLabs API~\footnote{\href{https://elevenlabs.io/}{elevenlabs.io}.}. Notably, all these models perform non-streaming ASR, as they require access to the full input sequence. We thus include a streaming variant of Whisper~\citep{whisper-streaming} (with a delay of 2.5s) and SeamlessStreaming~\citep{seamless}. We also include a (block) streaming version of Parakeet-TDT-v2, evaluated via running the official scripts 
with a 2.5s delay that matches the delay used by \oursASR{}.\footnote{We use chunk size of 10 frames (0.8s) and right context of 21 frames (1.7s).} 
For long-form transcription, we add the Distil-Whisper~\citep{distil-whisper} variant.

\looseness=-1
\textbf{Transcription quality.}\ We report micro-averaged Word Error Rate (WER), which %
avoids overemphasizing short sequences, and is the standard computation used in the OpenASR Leaderboard, of which we use the official evaluation codebase. %
Throughout this section, we use the Whisper normalizer for English~\citep{whisper}.

\looseness=-1
\textbf{Latency.}\ We evaluate the latency of streaming models as the average delay between the real timestamp of a word, and the time when this word is transcribed in the output. In the absence of ground-truth timestamps, we use pseudo-timestamps on Librispeech test-clean~\citep{librispeech} provided by~\cite{lugosch2019}. These timestamps were obtained by Montreal Forced Aligner~\citep{montreal_forced_aligner}, and use them as reference.%

\looseness=-1
\textbf{Timestamps.}\ See Appendix~\ref{sec:app:timestamps} for a complete description. 

\subsubsection{ASR Results}\label{sec:dsm_asr_results}
\looseness=-1
\textbf{Short-form transcription.}
Table~\ref{tab:openasr-leaderboard} shows that \oursASR{} is significantly better than streaming baselines, and even competitive with the best, non-streaming models of the OpenASR Leaderboard. With an average WER of 6.4\%---with 6.1\% being the current best score of the leaderboard---\oursASR{} is remarkably the only streaming model among top ASR systems. In Appendix~\ref{sec:app:timestamps}, we see that, in terms of timestamp precision, \oursASR{} performs significantly better than Whisper Large-v3 while somewhat underperforming CrisperWhisper, though with a better WER. 

\looseness=-1
\textbf{Long-form transcription.}
Table \ref{tab:longform-asr} reports WER values across 4 long-form datasets with sequences up to 2 hours: TED-LIUM~\citep{tedlium}, Meanwhile~\citep{whisper}, Rev16~\citep{whisper}, and Earnings21~\citep{earnings21}. %
The non-streamed version of Parakeet runs out of memory on a 80Gb GPU even when we use single-example batches, so we exclude it from the analysis.

From Table~\ref{tab:longform-asr}, we see that \oursASR{} outperforms streaming and non-streaming baselines except for Parakeet, which it closely matches (7.8 vs 7.9).

\textbf{Distillation}. At the pretraining stage, \oursASR{} is trained using pseudo-transcripts produced by \texttt{whisper-timestamped}~\citep{lintoai2023whispertimestamped}, which wraps Whisper Medium. As we can see from Table~\ref{tab:openasr-leaderboard}, the teacher model has the average WER of 8.1, while \oursASR{} gets 6.4. Without finetuning on datasets with ground-truth transcripts, \oursASR{} gets WER of 7.1.

We hypothesize that the observed improvements over the teacher come from (a) smoothing across a larger and more diverse set of real-world audio, implicitly leading to a domain adaptation, (b) low-temperature sampling that would eliminate non-systematic errors of the teacher model. Note that such improvements through teacher/student distillation have been observed before, e.g. on ImageNet classification \citep{yalniz2019billion}. We also perform augmentations such as codebook dropout.
After the fine-tuning stage, DSM-ASR gets WER of 6.4\%. Here, we train on ground-truth transcripts that come with the standard ASR datasets. At this stage, Whisper is only used to derive the timestamps.

\begin{figure}[t]
  \centering
  \begin{subfigure}{0.48\textwidth}
    \centering
    \includegraphics[width=0.8\linewidth]{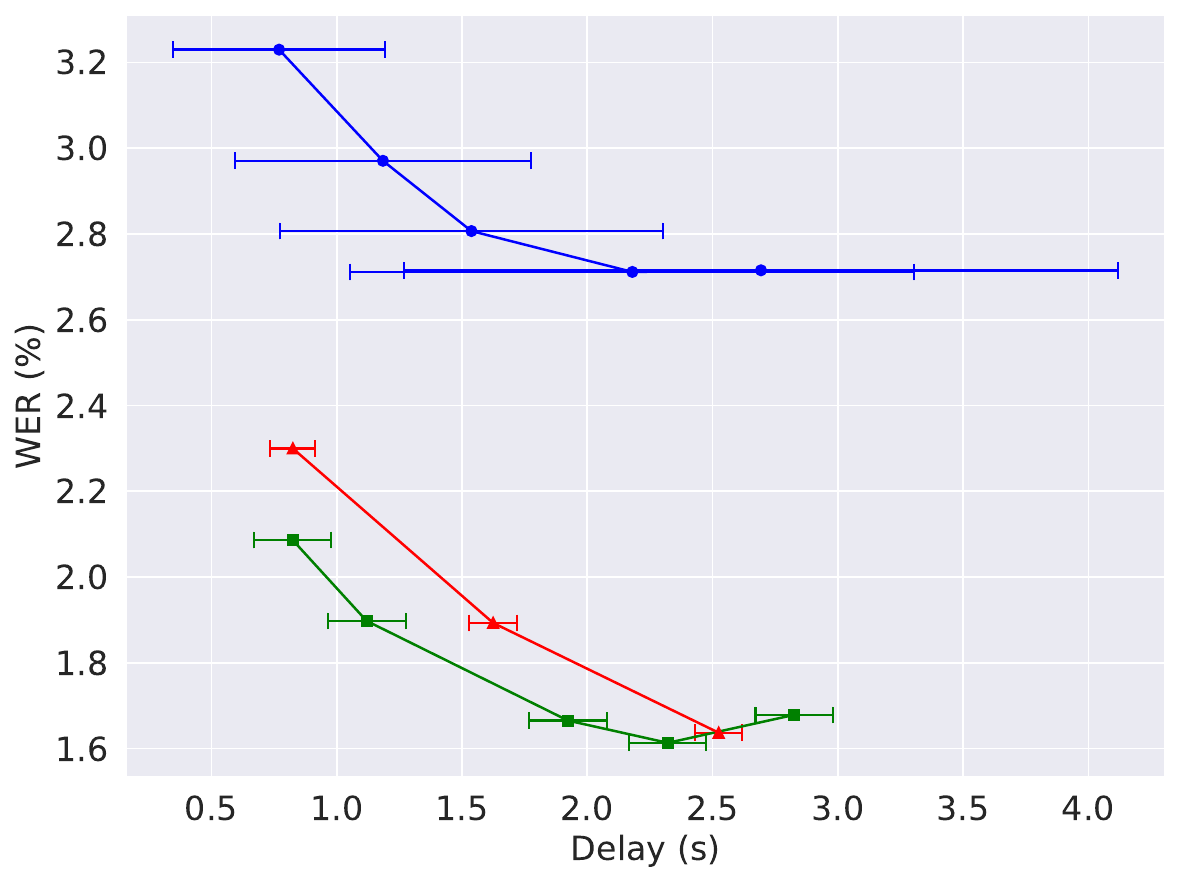}
    \label{fig:delay_vs_wer}
  \end{subfigure}
  \begin{subfigure}{0.48\textwidth}
    \centering
    \includegraphics[width=0.8\linewidth]{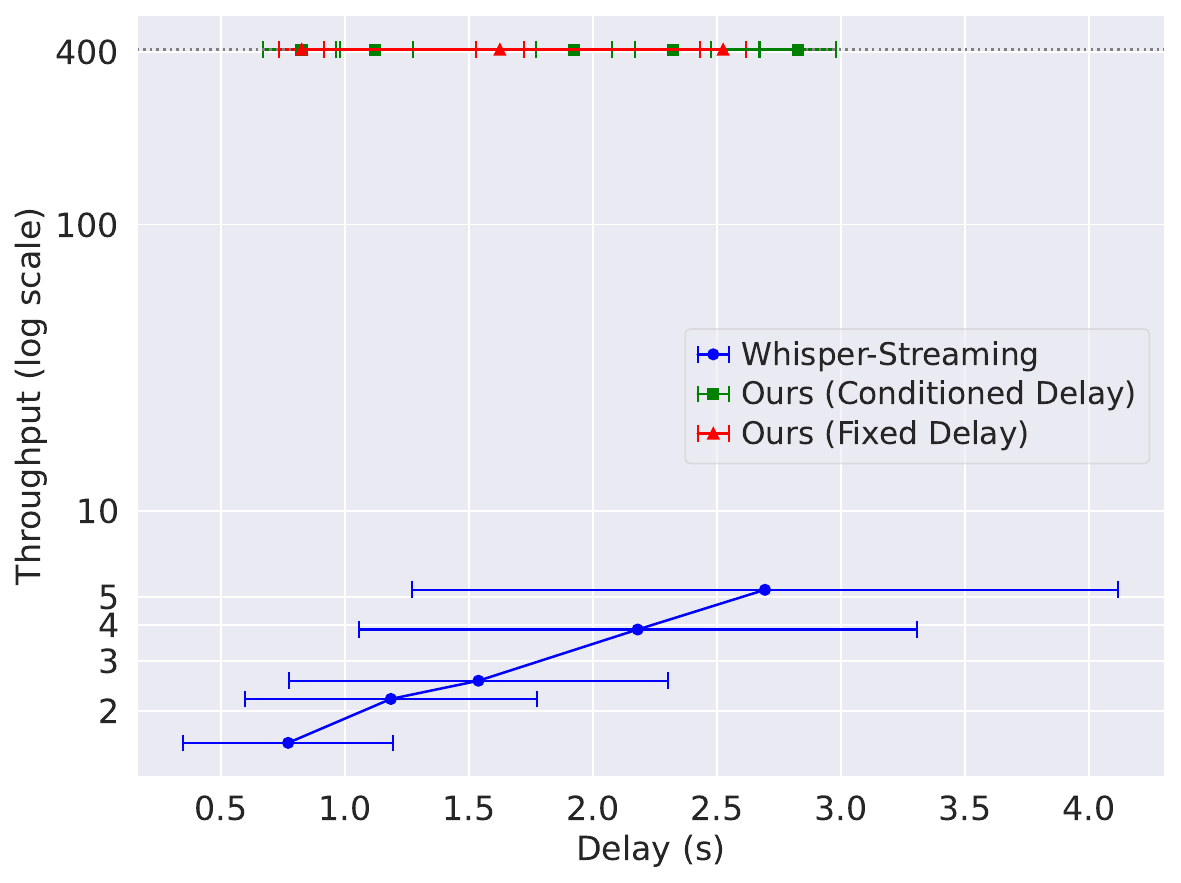}
    \label{fig:delay_vs_rtf}
  \end{subfigure}

  \caption{\textbf{ASR WER and throughput vs.\ delay}. Word Error Rate (WER, \%) (left) and throughput (right) in function of the delay. Throughput is the product between Real-Time Factor and batch size.%
  }
  \label{fig:delay}
\end{figure}

\looseness=-1
\textbf{Delay conditioning and latency.}
Figure~\ref{fig:delay} (left) compares the WER obtained for \oursASR{} with and without delay conditioning, along with Whisper-Streaming. We observe that the delay of Whisper-Streaming has a large variance, while \oursASR{} has a precision of $\sim$300ms around its target delay. 
Interestingly, training a single \oursASR{} model with delay conditioning outperforms
fixed delay variants.
Figure~\ref{fig:delay} (right) shows the throughput on an H100 GPU: \oursASR{} can process 400 sequences simultaneously while being real-time and its throughput is independent of the delay. This is unlike Whisper-Streaming which reduces its delay by re-evaluating the partial input sequence more frequently, increasing the computational cost. Combined with the fact that Whisper-Streaming does not allow for batching, this results in a 100x lower throughput than that of \oursASR{}.

\begin{table}[t]
\tiny
\centering
\caption{\textbf{Long-form TTS WER}. We compare open-source and closed-source baselines. 
Short-form inference is ran with a short context. ElevenLabs is evaluated on a limited subset due to its cost.}
\label{tab:longform}
\begin{sc}
\begin{tabular}{l|r|rr|r|rr}
\toprule
  &  \multicolumn{3}{c|}{WER English (\%) ($\downarrow$)}  &
    \multicolumn{3}{c}{WER French (\%) ($\downarrow$)}  \\
\textbf{Model} & \textbf{Avg.} & \textbf{Dialogs} & \textbf{monologues} & \textbf{Avg.} & \textbf{Dialogs} & \textbf{monologues}\\
\midrule
\multicolumn{7}{c}{\textit{Short-form inference}} \\
\midrule[0.01pt]
orpheus & 3.51 & 3.91 & 3.11 & - & - & - \\
csm & 3.82 & 2.95 & 4.68 & - & - & - \\
dia & 2.79 & 2.40 & 3.18 & 14.20 & 12.49 & 15.91\\
chatterbox & 1.95 & 1.43 & 2.47 & - & - & - \\
\oursTTS{} (ours, turn-by-turn) & 2.06 & 1.36 & 2.75 & 3.20 & 2.74 & 3.66\\
\midrule
\multicolumn{7}{c}{\textit{Long-form inference}} \\
\midrule[0.01pt]
\oursTTS {} (ours) & \textbf{1.72} & \textbf{1.15} & \textbf{2.29} & \textbf{2.96} & \textbf{2.66} & \textbf{3.26}\\
\midrule
\multicolumn{7}{c}{\textit{Long-form inference, small subset}} \\
\midrule[0.01pt]
\oursTTS{} (ours) & 2.35 & 1.76 & \textbf{2.94} & \textbf{2.78} & \textbf{2.91} & \textbf{2.65}\\
ElevenLabs Flash & 2.59 & 1.13 & 4.05 & 4.51 & 4.38 & 4.64\\
ElevenLabs Multilingual v2 & \textbf{2.01} & \textbf{0.91} & 3.10 & 2.93 & 3.12 & 2.75\\
\bottomrule
\end{tabular}
\end{sc}
\end{table}

\begin{table}[t]
\tiny
\centering
\caption{\textbf{Long-form TTS Speaker Similarity}. We compare open-source and closed-source baselines. 
Short-form inference is ran with a short context. ElevenLabs runs on a limited subset due to its cost.}
\label{tab:longform_similarity}
\begin{sc}
\begin{tabular}{l|r|rr|r|rr}
\toprule
  &  \multicolumn{3}{c|}{Speaker Sim. English (\%) ($\uparrow$)}  &
    \multicolumn{3}{c}{Speaker Sim. French (\%) ($\uparrow$)}  \\
\textbf{Model} & \textbf{Avg.} & \textbf{Dialogs} & \textbf{monologues} & \textbf{Avg.} & \textbf{Dialogs} & \textbf{monologues}\\
\midrule
\multicolumn{7}{c}{\textit{Short-form inference}} \\
\midrule[0.01pt]
orpheus & 38.86 & 36.80 & 40.92 & - & - & -\\
csm & 74.44 & 65.83 & 83.05 & - & - & -\\
dia & 62.48 & 57.03 & 67.93 & 54.99 & 50.13 & 59.85\\
chatterbox & 66.92 & 63.32 & 70.51 & - & - & -\\
\oursTTS{} (ours, turn-by-turn) & \textbf{80.90} & \textbf{77.67} & \textbf{84.13} & \textbf{80.46} & \textbf{76.63} & \textbf{84.29}\\
\midrule
\multicolumn{7}{c}{\textit{Long-form inference}} \\
\midrule[0.01pt]
\oursTTS{} (ours) & 76.46 & 74.06 & 78.87 & 75.96 & 73.21 & 78.72\\
\midrule
\multicolumn{7}{c}{\textit{Long-form inference, small subset}} \\
\midrule[0.01pt]
\oursTTS{} (ours) & \textbf{76.40} & \textbf{73.43} & \textbf{79.37} & \textbf{75.92} & \textbf{72.98} & \textbf{78.86}\\
ElevenLabs Flash & 46.84 & 45.39 & 48.30 & 53.55 & 53.97 & 53.13\\
ElevenLabs Multilingual v2 & 56.40 & 55.36 & 57.44 & 62.10 & 60.40 & 63.79\\
\bottomrule
\end{tabular}
\end{sc}
\end{table}

\subsection{Text-To-Speech experiments}
\label{sec:experiments:tts}

\textbf{Evaluation datasets.} We collect a novel dataset for long-form TTS evaluation in English and French. We first use news articles from the NTREX-128~\citep{federmann-etal-2022-ntrex} text dataset,
given 123 monologues per language. To evaluate controllable dialog capabilities, we use 110 synthetic scripts per language generated by an LLM, spanning three categories: daily life, technical, and number-heavy discussions. For voice conditioning, we use samples from the test set of VCTK~\citep{Yamagishi2019CSTRVC} for English, and from the test and valid sets of CML~\citep{Cmltts2023} for French. We provide examples and more details in the Appendix~\ref{sec:app:eval_data}, while the dataset and evaluation code is available at \href{https://github.com/kyutai-labs/tts_longeval}{github.com/kyutai-labs/tts\_longeval}.

\looseness=-1
\textbf{Metrics.} We evaluate the per-document WER, using text normalization from Whisper~\citep{whisper}.
We collect subjective metrics covering both the speaker similarity to the conditioning and overall speech quality, see Appendix~\ref{sec:app:data} for more details. 

\textbf{Baselines.} We compare to open-source models Chatterbox, Dia, Orpheus, and CSM, as well as ElevenLabs.\footnote{
\href{https://github.com/resemble-ai/chatterbox}{resemble-ai/chatterbox},
\href{https://github.com/nari-labs/dia}{nari-labs/dia},
\href{https://github.com/canopyai/Orpheus-TTS}{canopyai/Orpheus-TTS},
\href{https://github.com/SesameAILabs/csm}{SesameAILabs/csm} and \href{https://elevenlabs.io/}{elevenlabs.io}.} Dia and ElevenLabs support French and English, while Chatterbox, Orpheus and CSM only support English\footnote{A new version of Chatterbox supports multiple languages but was not available at the time of this study.}. Chatterbox, Dia, Orpheus and CSM can be speaker-conditioned through prefixing, with Dia and CSM supporting dialogs. For Chatterbox, Orpheus and ElevenLabs, dialogs are emulated by concatenating single-speaker turns. Details of how baselines are evaluated are provided in Appendix~\ref{sec:app:tts_baselines}.

\subsubsection{TTS Results}

\begin{table}[t]
\centering
\tiny
\caption{\textbf{Subjective evaluations on TTS}. We compare with baselines
over two axes through human evaluations: speech quality, measured as MUSHRA scores (1--100, along with std. of the mean), and speaker-similarity win-rates, summarized as ELO scores (with 95\% confidence intervals).}
\label{tab:subjective_tts}
\begin{sc}
\begin{tabular}{l|rr|rr}
\toprule
  &  \multicolumn{2}{c|}{English} &
    \multicolumn{2}{c}{French} \\
\textbf{Model} & \textbf{Quality} ($\uparrow$)  & \textbf{Spk.\ Sim.} ($\uparrow$)  & \textbf{Quality}($\uparrow$)  & \textbf{Spk.\ Sim.}($\uparrow$)  \\
\midrule
Orpheus & $27.1 \pm 2.1$ & $1716.3 \pm 35.6$ & - & -\\
CSM & $39.6 \pm 2.2$ & $1997.6 \pm 19.4$ & - & -\\
Dia & $43.6 \pm 2.4$ & $1926.9 \pm 21.3$ & $24.9 \pm 2.2$ & $1679.0 \pm 41.7$\\
Chatterbox & $\mathbf{67.7} \pm 2.0$ & $2068.3 \pm 18.7$ & - & -\\
DSM-TTS (ours) & $54.8 \pm 2.3$ & $2048.3 \pm 18.6$ & $62.5 \pm 2.0$ & $2098.5 \pm 21.1$\\
DSM-TTS (ours, turn-by-turn) & $53.1 \pm 2.2$ & $\mathbf{2092.6} \pm 18.3$ & $61.9 \pm 2.0$ & $\mathbf{2179.1} \pm 22.4$\\
ElevenLabs Flash & $65.0 \pm 2.2$ & $1932.3 \pm 20.9$ & $61.5 \pm 2.3$ & $1894.3 \pm 20.4$\\
ElevenLabs Multilingual v2 & $59.7 \pm 2.4$ & $2035.3 \pm 19.2$ & $\mathbf{66.4} \pm 2.0$ & $2047.3 \pm 21.1$\\

\bottomrule
\end{tabular}
\end{sc}
\end{table}

\begin{table}[t]
\centering
\tiny
\caption{\textbf{TTS: Latency and throughput}.
We compare the inference performance of \oursTTS{} and selected baselines.
Real-Time Factor (RTF) is higher than 1 if the model can produce audio in real time.
Throughput is the product between Real-Time-factor and batch size.
}
\label{tab:latency}
\begin{sc}
\begin{tabular}{l|r|rrr}
\toprule
\textbf{Model} & \textbf{Model Size} & \textbf{Latency} (ms)($\downarrow$)  & \textbf{RTF}($\uparrow$)  & \textbf{Throughput}($\uparrow$) \\
\midrule
Dia & 1.6B & - &  0.7 & 0.7 \\
CSM & 1.5B &- & 1.0 & 1.0 \\
Orpheus & 3.8B & - & 0.7 & 0.7 \\
Chatterbox & 0.8B & - & 1.8 & 1.8 \\
\oursTTS{} b.s.=1  & 1.8B & 150 & 3.2 & 3.2 \\
\oursTTS{} b.s.=32 & 1.8B & 380 & 2.4 & 76.8 \\
\oursTTS{} b.s.=64 & 1.8B & 403 & 2.1 & 137.3 \\
\bottomrule
\end{tabular}
\end{sc}
\end{table}

\textbf{Main results.} As seen in Table~\ref{tab:longform}, our approach provides the
lowest WER across all languages for both monologues and dialogs. Our method is the only one to run long-form inference across all cases, CSM showing strong degradation when running with longer sequences, Dia only being trained for 20s output, and ElevenLabs requiring per-turn generation for dialogs. We also provide speaker similarity measurements in Table~\ref{tab:longform_similarity}.
We report subjective results in Table~\ref{tab:subjective_tts}.
When evaluating turn-by-turn (e.g. short-form like the other baselines), we outperform all existing methods in terms of speaker similarity, while still surpassing commercial methods when using long-form generation.
In terms of quality, Chatterbox is rated the highest, with \oursTTS{} surpassing the other open source baselines. 
Note that we kept all methods with their original sample rate (e.g.\ 44.1kHz for ElevenLabs) which can contribute to the difference.

\textbf{Throughput and latency.} Our method is easily batchable, leading to gains in throughput while staying compatible with real-time generation. As shown in Table~\ref{tab:latency}, on a single H100 the amount of audio generated is 100x real-time. More details are provided in Appendix~\ref{sec:app:latency}.

\begin{table}
\centering
\caption{\textbf{Results of \oursTTS{} trained on public datasets, evaluated on LibriSpeech test-clean}. We compare to F5-TTS~\citep{chen2024f5}, following their evaluation framework. RTF and Throughput were computed on a H100 for both methods.}
\label{tab:tts-libri}
\resizebox{\textwidth}{!}{
\begin{tabular}{lcc|rr|rrr}
\toprule
\textbf{Model} & \textbf{\#Param.} & \textbf{\#Data}
& \textbf{WER} (\%) $\downarrow$ & \textbf{SIM-o} $\uparrow$ & \textbf{Latency} $\uparrow$ & \textbf{RTF} $\uparrow$ & \textbf{Throughput} $\uparrow$
\\
\midrule
F5-TTS (16 flow steps) & 336M & 100K (EN, ZH) & 2.53 & 0.66 & $\sim 800\,\text{ms}$ & \textbf{14.2} & 14.2 \\
F5-TTS (32 flow steps) &  & & 2.42 & 0.66 &  $\sim 400\,\text{ms}$  & 7.1 & 7.1 \\
\midrule[0.01pt]
\oursTTS{} (ours, nq=16, b.s.=1) & 750M & 88K (EN) & 1.95 & 0.67 & \textbf{139ms} & 4.4 & 4.4\\
\oursTTS{} (ours, nq=32, b.s.=1) & 900M & 88K (EN) & \textbf{1.68} & \textbf{0.71} & 172ms & 2.7& 2.7\\
\oursTTS{} (ours, nq=32, b.s.=32) & 900M & 88K (EN) & \textbf{1.68} & \textbf{0.71} & 351ms & 2.3 & \textbf{74.2}\\
\bottomrule
\end{tabular}
}
\end{table}

\begin{table}[t]
\centering
\caption{\textbf{Results of \oursTTS{} trained on public datasets, evaluated on Seed test en}. See Section~\ref{app:sec:libri_seed} for details.}
\label{tab:tts-seed}
\begin{tabular}{l|rr}
\toprule
\textbf{Model} 
& \textbf{WER} (\%) $\downarrow$ & \textbf{SIM-o} $\uparrow$
\\
\midrule
F5-TTS~\citep{chen2024f5}  & 1.83 & \textbf{0.71} \\
Cosyvoice 3-1.5B (RL)~\citep{du2025cosyvoice}  & \textbf{1.45} & 0.70 \\
\oursTTS{} (ours, nq=16) & 1.58 & 0.70 \\
\oursTTS{} (ours, nq=32) & 1.71 & \textbf{0.71} \\
\bottomrule
\end{tabular}
\end{table}

\textbf{Training on public data only.}
For reproducibility, we provide results training from scratch \oursTTS{} with the public training data described in Section~\ref{app:sec:libri_seed}, totaling 88k hours of speech.
We train a 300M parameters backbone model with either 16 or 32 codebooks. We evaluate
on Librispeech test clean~\citep{librispeech} and on Seed test en~\citep{anastassiou2024seed}. We provide the results in Tables~\ref{tab:tts-libri} and \ref{tab:tts-seed}, with comparisons to F5-TTS~\citep{chen2024f5} and CosyVoice3-1.5B (RL) ~\citep{du2025cosyvoice}.
Refer to Appendix~\ref{app:sec:libri_seed} for details.

\textbf{Further results and ablations.}  Finally, we provide ablations on the codec choice, and the contribution from the action and lookahead stream in the Appendix~\ref{app:sec:ablations}.

\textbf{Watermarking.} A number of open source baselines in Table~\ref{tab:longform} rely on watermarking to protect against negative use of their models. Note that for open source models, this step can be disabled. Besides, we noticed that a single round of encoding and decoding with the Mimi codec with 32 codebooks completely or largely remove the watermarks commonly used, see Appendix~\ref{app:sec:watermark}. This suggests that securing the use and traceability of modern TTS models is still an open question.  

\textbf{\oursASR{} and \oursTTS{} as a speech interface for LLMs.} We combine \oursASR{}, \oursTTS{}, and Gemma 3~\citep{gemmateam2025gemma3technicalreport} into an LLM voice chat application with sub-second latency. The application is available at \url{https://unmute.sh}.

\section{Conclusion}
\looseness=-1
We introduce Delayed Streams Modeling, a flexible framework for streaming sequence-to-sequence learning. DSM provides a remarkable trade-off between quality and latency, and an unprecedented throughput among streaming models. Focusing on speech-text tasks, \oursASR{} is the first streaming ASR model to provide timestamped, formatted transcripts that competes with the top offline models, while \oursTTS{} is competitive with non-streaming baselines while being the only model providing long form synthesis. In future work, we will extend DSM to more sequential multimodal tasks. 
In particular, one limitation of our approach is the need for aligned domains, which reduces the amount of gold-standard ground-truth data that can be used for training. 

\looseness=-1
\textbf{Societal impact.} We acknowledge that streaming naturalistic speech with voice conditioning opens up both opportunities in inclusive human-machine interactions and risks of fraudulent impersonation. Addressing the latter requires that public access to such technologies is accompanied by proper user terms, voice verification mechanisms 
, and resilient watermarking of generated content. Given the limitations of such existing approaches, in particular for open source models, we have not open sourced the voice conditioning module for our best TTS model, only providing pre-computed speaker embeddings.

\clearpage
\bibliographystyle{icml2025}
\bibliography{references}
\clearpage
\input{80_appendix}
\end{document}

%% file: 05_declarations.tex
\usepackage[utf8]{inputenc} %
\usepackage[T1]{fontenc}    %
\usepackage{hyperref}       %
\usepackage{url}            %
\usepackage{booktabs}       %
\usepackage{amsfonts}       %
\usepackage{nicefrac}       %
\usepackage{microtype}      %
\usepackage[dvipsnames]{xcolor}         %
\usepackage{amsmath}
\usepackage{amssymb}
\usepackage{stmaryrd}
\usepackage{mathtools}
\usepackage{subcaption} %
\usepackage{tcolorbox}

\newcommand{\TODO}[1][]{\textcolor{red}{\bf [TODO]}}

\newcommand{\oursASR}[1][]{DSM-ASR}
\newcommand{\oursTTS}[1][]{DSM-TTS}
\newcommand{\ours}[1][]{DSM}

\newcommand{\proba}[1]{\mathbb{P}\left[#1\right]}
\newcommand{\reals}{\mathbb{R}}
\newcommand{\tY}{\tilde{Y}}
\newcommand{\qalo}[1]{q_{\text{aligned}}}
\newcommand{\qal}[1]{\qalo{}\left(#1\right)}
\newcommand{\qdelo}[1]{q_\tau}
\newcommand{\qdel}[1]{\qdelo{}\left(#1\right)}
\newcommand{\nats}{\mathbb{N}}
\newcommand{\spectoken}[1]{\textsc{\ttfamily{}#1}}

\usepackage{enumitem} %

\usepackage{graphicx} %

%% file: 30_method.tex
\begin{figure}[t]
\centering
\includegraphics[width=.7\textwidth]{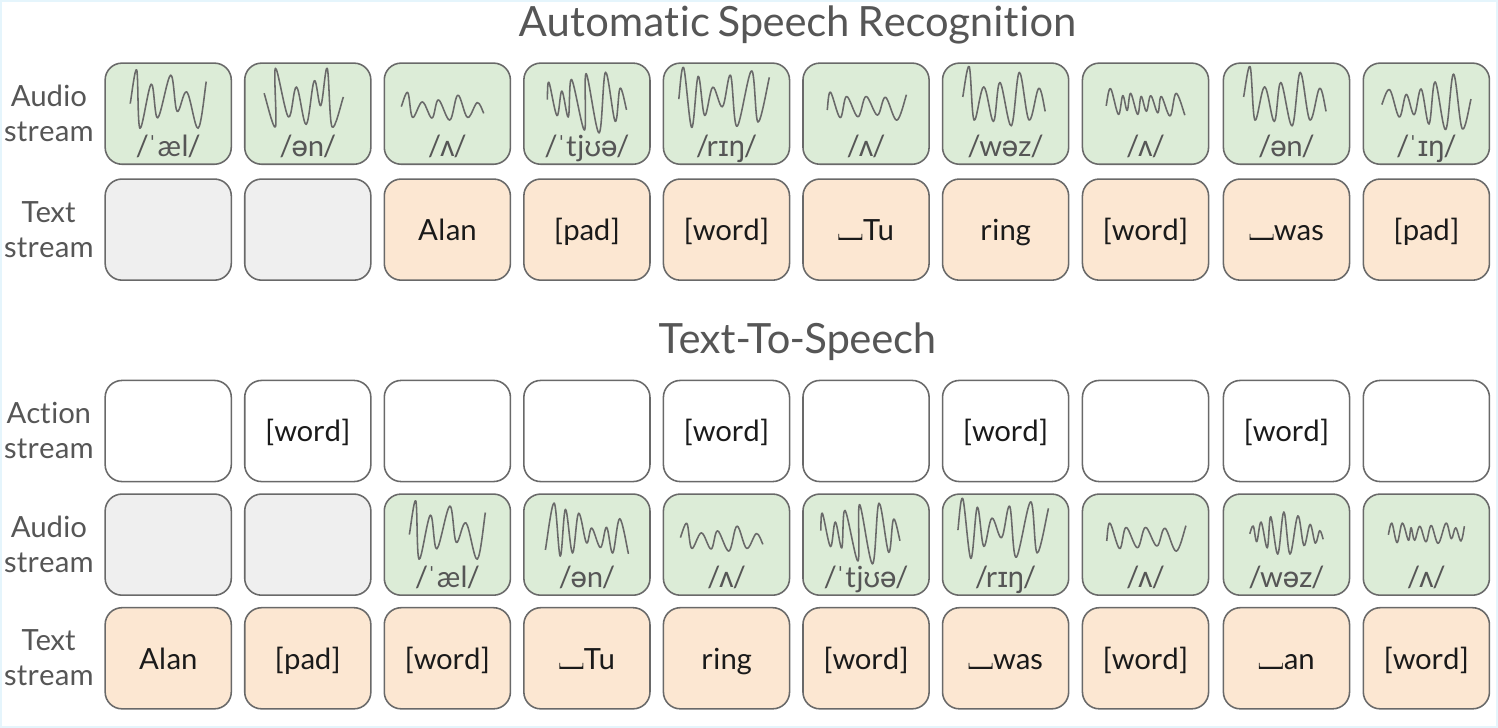}
\caption{\textbf{Delayed streams modeling for speech-text tasks}. Depending on which stream is delayed with respect to the other, we solve either an ASR or a TTS task. For TTS, we further need an action stream for the model to let us know when it is ready to receive a new word.
}
\label{fig:concept}
\end{figure}

\section{Method}

\label{sec:modeling}
\textbf{Notation.}\ 
We wish to solve a sequence-to-sequence task between two domains 
$\mathcal{X}$ and $\mathcal{Y}$. Each domain consists of sequences of vectors of all possible lengths, e.g. 
\begin{equation}
\label{eq:domain_def}
    \mathcal{X} = \bigcup_{T \in \nats} \{ (X_{t}) \in \reals^{T \times d }\}, \qquad
    \mathcal{Y} = \bigcup_{T' \in \nats} \{ (Y_{t'}) \in \reals^{T' \times d' }\}.
\end{equation}
In the case where either $X_t$ or $Y_t$  is discrete-valued, we can use a one-hot representation for it in Eq.~\eqref{eq:domain_def}.
We assume that we are given a joint probability distribution over the outer product domain $\mathcal{X}\times \mathcal{Y}$, and that we have
the random variables $X \in \mathcal{X}$ and $Y \in \mathcal{Y}$, along with the joint distribution \begin{equation}\proba{X, Y} = p(X ,Y). \end{equation}
We also introduce $T \in \nats$ (resp.\ $T'$) the random variable indicating
the length of $X$ (resp.\ $Y$), along with the marginals $p(X)$ and $p(Y)$. For any sequence $Z$, and index $t$, we denote $Z_{<t} = (Z_1, \ldots, Z_{t-1})$, %
potentially empty if $t \leq 0$. We similarly define $Z_{\leq t}$, $Z_{\geq t}$, and $Z_{> t}$.

\textbf{Sequence-to-sequence as joint modeling.}\ 
Let's assume for this paragraph that $\mathcal{X}$ is the set of all possible monophonic waveforms sampled at 24 kHz,
and $\mathcal{Y}$ is made of sequences of one-hot encoded vectors over a set of words.
Intuitively, we assume there exists a coupling $p(X, Y)$ such that $p(X, Y)$ is high
if $Y$ represents the transcription of $X$, or conversely, if $X$ represents a speech utterance
of the text given by $Y$.
Formally, the task of ASR corresponds to sampling from the distribution $\proba{Y | X}$, 
while the task of TTS corresponds to sampling from the distribution $\proba{X | Y}$.
Thus, each task can be solved by accurately estimating both probability distributions,
\begin{equation}
    \label{eq:estimate}
    q(X, Y)\approx \proba{Y | X}, \qquad q'(Y, X) \approx \proba{X | Y}.
\end{equation}

For simplicity, we now only focus on estimating $\proba{Y|X}$, the inverse task being obtained by
exchanging the definition of $X$ and $Y$. We thus call $\mathcal{X}$ the input domain, and $\mathcal{Y}$ the output domain.

\textbf{Auto-regressive modeling of $Y$.}\ 
A good candidate for estimating $\proba{Y | X}$ is auto-regressive modeling, with a Transformer model~\citep{attentionvaswani}, under the extra assumption that the output domain $\mathcal{Y}$ can be discretized. Thus, one would estimate
\begin{equation}
    q(y | X, Y_{< t}) \approx \proba{Y_t=y | X, Y_{<t}}.
\end{equation}

One can then sample $Y$ auto-regressively, knowing $X$.
Due to the lack of explicit structure between the time grid $t$ of $X$ and $t'$ of $Y$,
one would usually condition on the entirety of $X$, e.g.\ when using Transformer based models, 
either by prefixing the entire sequence
$X$ before the generation $Y$, or by providing $X$ through cross-attention layers, which is mathematically equivalent. 
This forbids the use of the model in a streaming fashion, as the entire input signal $X$ must be known ahead of time, and cannot be extended once the generation of $Y$ has started.
Such methods often require explicit and manual chunking and stitching operations, which also reduces their ability to be efficiently batched. Conversely, aligning $X$ and $Y$ to the same frame rate allows for batched streaming inference.

\textbf{Aligning sequences for streaming prediction.}
We assume that both domains $\mathcal{X}$ and $\mathcal{Y}$ can share the same time grid,
e.g.\ $(X_t)\in\reals^{T \times d}$ and $(Y_t)\in\reals^{T \times d'}$. 
We call two such aligned sequences \emph{streams}. 
Then one can simply model%
\begin{equation}
\label{eq:q_aligned}
    \qal{y | X_{\leq t}, Y_{< t}} \approx \proba{ Y_t = y|X_{\leq t}, Y_{< t}}.
\end{equation}
Given $X\sim p(X)$, we sample auto-regressively from Eq.~\eqref{eq:q_aligned}, \emph{with a streaming context} $X$,
\begin{equation}
\label{eq:sample_ty}
   \tY_1 \sim  \qal{\tY_1 | X_1}, \qquad
   \tY_t \sim \qal{\tY_t | X_{\leq t}, \tY_{< t}}.
\end{equation}
We would want that given $X\sim p(X)$, then $(X, \tY)\sim(X, Y)$,
so that in particular $\proba{\tY | X} \approx \proba{Y | X}$.
However this needs not be the case unless certain conditions are met.

\textbf{The importance of causality.}\ 
In particular, for $(X, \tY)\sim(X, Y)$ to be true,
$Y_{> t}$ must be independent of $X_{> t}$, knowing $X_{\leq t}$.
To realize that, one can look at a simple counter-example taking
$X_t \sim \mathcal{B}(0.5)$ independent Bernoulli variables,
and $Y_t = X_t \oplus X_{t+1}$ the XOR of $X_t$ and $X_{t+1}$.
Clearly $\proba{Y_t | X_{\leq t}, Y_{< t}} \sim \mathcal{B}(0.5)$ for all $t$, yet, given $X = (0, 1)$, one would have 
\begin{equation*}
Y_1 = 1 \quad\text{a.s.},\qquad \tY_1 \sim \mathcal{B}(0.5).
\end{equation*}
Thus $Y | X$ and $\tY | X$ have different distributions. Intuitively, given that we do not sample $X$ but teacher-force real-world data, we must ensure that when sampling $\tY_t$, no future value of $X_{>t}$ might end up in ``contradiction'' with the value we sampled.

\textbf{Delaying the output stream.}\ 
In practice, this is achieved by delaying the output stream $Y_t$ by a number of steps $\tau > 0$.
Thus, we replace Eq.~\eqref{eq:q_aligned} by
\begin{equation}
\label{eq:q_delayed}
\qdel{y | X_{\leq t + \tau}, Y_{< t}} \approx \proba{ Y_t=y |X_{\leq t +\tau}, Y_{< t}},
\end{equation}
and define $\tY^\tau$, similarly to the procedure described in Eq.~\ref{eq:sample_ty}.
Perfect independence is hard to achieve: in the case of ASR, a named entity might be ambiguous without context, and only future development in a discussion would resolve this ambiguity. Taking $\tau = T$ recovers the prefixing or cross-attention approaches presented earlier. In practice, there is a trade-off between the level of independence of $Y_t$ with $X_{> t + \tau}$, and the latency of the method.

\begin{figure}
\centering
\begin{minipage}[t]{.42\textwidth}
\centering
\includegraphics[width=0.9\textwidth]{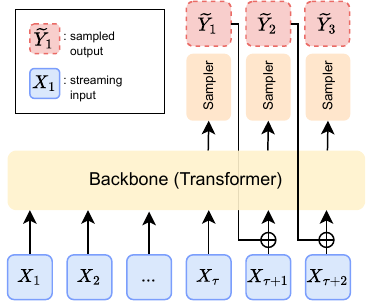}
\caption{\textbf{DSM Architecture}. %
\looseness=-1
Transformer is fed %
with the streaming input $X_t$. After a delay $\tau$, a sampler is fed with the output of the backbone samples $\tY_t$. At the next step, the backbone receives both the sampled value and next streaming input, whose embeddings are summed.}
\label{fig:architecture}
\end{minipage}\hspace{.03\textwidth}%
\begin{minipage}[t]{0.55\textwidth}
\centering
\includegraphics[width=0.9\textwidth]{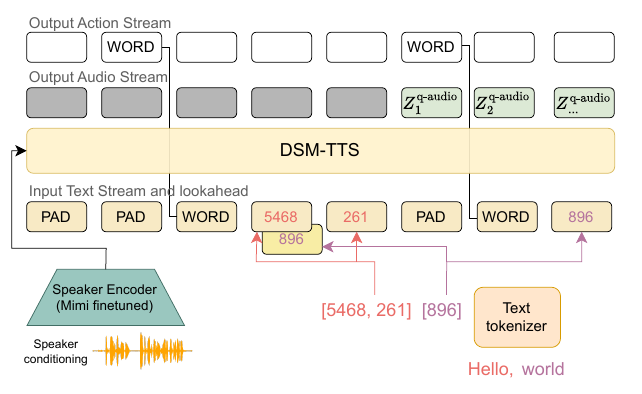}
\caption{\textbf{DSM-TTS inference}.
The input words ``{\color{BrickRed}Hello,} {\color{Orchid}world}'' are tokenized. Until the model action stream outputs a \spectoken{WORD}, it is fed with \spectoken{PAD}. Then the first word's tokens are fed, including a look-ahead text stream. Once a delay $\tau$ (here 5 for illustration purposes) has accumulated, the model also outputs the audio.}
\label{fig:tts_inference}
\end{minipage}
\end{figure}

\looseness=-1
\textbf{Architecture.}\ DSM, depicted in Figure~\ref{fig:architecture}, contains three components:
(i) an auto-regressive backbone, (ii) an input embedder for $X$ and $Y$ into the backbone, and (iii) a sampler for $\tY^\tau$ conditioned on the output of the backbone. 
The backbone can be a Transformer architecture, optionally equipped with cross-attention layers
to provide further non-streaming contextual information.
The embedder for $X$ and $Y$ can be learnt embedding tables in the case where both domains are discrete. The embeddings are summed before going into the backbone. On the output side, we mask the loss on the tokens of $X$ and only compute cross-entropy on $Y$. Finally, the conditional sampler can be a linear layer applied to the output of the backbone to derive logits if $Y$ is discrete. It could also be a flow or diffusion model conditioned on the output of the backbone for the continuous case.

\subsection{Representations of the speech and text domains}
\label{sec:representations}
We demonstrate the DSM framework on ASR and TTS, where the two domains are text and audio. %

\looseness=-1
\textbf{Audio.}\ 
Given a waveform $w \in \reals^{d_s \cdot f_s}$ with the duration in seconds $d_s$ and the sample rate $f_s=24\,\text{kHz}$, we turn it into a more compact latent space using the Mimi codec~\citep{moshi}, giving us
a sequence of tensors $Z^\text{audio} \in \reals^{d_s \cdot f_r \times d_{\text{audio}}}$, 
with a frame rate of $f_r=12.5\,\text{Hz}$.
This latent space is discretized with Residual Vector Quantization~\citep{soundstream} (RVQ), giving us
a set of $Q \in \llbracket 1, 32\rrbracket$ coarse-to-fine discrete values per time step with cardinality $N_a{=}2048$, each coming from one codebook in the RVQ, giving a quantized representation $Z^\text{q-audio} \in \{1, \ldots, N_a\}^{d_s \cdot f_r \times Q}$.

\looseness=-1
\textbf{Text.}\ 
We tokenize text using a vocabulary of $N_t$, specifically trained on speech data transcriptions.
Two tokens have a special meaning: \spectoken{PAD} (indicating the absence of words at this time) and \spectoken{WORD} (indicating the start of a new word following~\citet{moshi}.
Given a transcript, with word-level timestamps, of a waveform of duration $d_s$, its aligned text representation is $Z^\text{text}\in\{1, \ldots N_t\}^{d_s \cdot f_r}$.
For each word in the transcript represented by tokens $(x_1, \ldots, x_n) \in \{1, \ldots, N_t\}^n$ and starting at $s \in \reals^+$ seconds, we
define its start index $i = \mathrm{floor}(s \cdot f_r)$, and store it as $Z^\text{text}_{i} \leftarrow \spectoken{WORD}$, 
$Z^\text{text}_{i+1}\leftarrow x_1$, $Z^\text{text}_{i+2}\leftarrow x_2$, etc. Any step in $Z^\text{text}$ not assigned by a word token is given the special value \spectoken{PAD}. 

\subsection{DSM for automatic speech recognition: DSM-ASR}
\label{sec:dsm_asr}

\looseness=-1
For ASR, we consider $X = Z^\text{q-audio}$  and $Y = Z^\text{text}$. By predicting the word tokens of $Y$, we learn to transcribe audio, while computing the loss on \spectoken{PAD} and \spectoken{WORD} tokens trains the model to predict the precise boundaries of each word. At inference time, we teacher-force the audio tokens of $X$ and sample the full sequence $Z^\text{text}$ to obtain a transcription along with timestamps with a precision of 80ms (frame size). %
This is allowed by the fact that we apply a constant delay to all words in the sequence, meaning we only need to shift the output timestamps back by the same value to recover the true timestamps. %

\looseness=-1
\textbf{Deriving aligned speech-text data.}\ 
We are looking for fine-grained alignment between speech and text, however speech datasets are typically aligned at the level of the sentence~\citep{librispeech}. Conveniently, \texttt{whisper-timestamped}~\citep{lintoai2023whispertimestamped} provides automatic transcriptions with word-level timestamps. We rely on these pseudo-labels for the pretraining phase of \oursASR{}. We then finetune on a mixture of public datasets with ground-truth transcripts (see details in Section ~\ref{sec:training_protocol}), which pose two challenges.
First, the automatic transcriptions extracted by Whisper in pretraining are formatted with capitalization and punctuation, but the level of formatting varies a lot between datasets.
To address this, we train a 300M prefix-LM for automatic formatting, on a dataset of formatted Whisper transcripts. %
A second challenge is that these ground-truth transcripts do not have word-level alignment. We derive those by producing pseudo-transcripts with Whisper, and reconciling them with the formatted transcript using a Dynamic Time Warping algorithm~\citep{giorgino2009computing}.

\textbf{Delay conditioning for inference-time control.}
\looseness=-1
As shown in Section~\ref{sec:dsm_asr_results},  transcription quality is heavily dependent on the delay between audio and text. Thus, training \oursASR{} with a fixed delay requires choosing a latency/quality trade-off beforehand, and retraining a new model for each delay, despite the training task remaining fundamentally the same. To instead control this trade-off at inference, we train \oursASR{} over random delays, sampled for each sequence. %
The model is additionally conditioned on a cosine embedding~\citep{attentionvaswani} of the delay (expressed in milliseconds), added to the inputs. Experiments in Section~\ref{sec:dsm_asr_results} compare this model to the models trained with a fixed delay %
and show that the effective delay precisely respects the conditioning value.

\subsection{DSM for text-to-speech}
\label{sec:dsm_tts}
\looseness=-1
We further apply DSM to TTS, taking $X = Z^\text{text}$, $Y = Z^\text{q-audio}$. We use a stream delay of 1.28s (or 16 steps) on the output audio.
For sampling along the $Q$ dimension in $Z^\text{q-audio}$, we use a RQ-Transformer as a sampler~\citep{rq-transformer, moshi}, i.e.\ a smaller Transformer conditioned on the output of the backbone at each timestep and performing autoregressive modeling along the $Q$ dimension.
All the backbone inputs (generated audio tokens and next word token input) are fed through learnt embeddings and summed.
We are confronted with the problem that the input domain is no longer plain text, but text properly padded for time alignment.
While at train time we can teacher-force the ground-truth padded text, this is not the case for a novel text to synthesize at inference time.

\textbf{Action output stream.}
\looseness=-1
We add an extra stream to the TTS outputs,
whose goal is to predict whether the next input text token will be a \spectoken{WORD} token or not. This special input token indicates that a new word is starting, and that its tokens are going to follow as inputs.
This extra stream controls an inference-time \emph{action}: when predicted by the model, we will feed as input the text tokens for the next word over the next time steps.
While these are being fed, the model is not allowed to output another \spectoken{WORD} action. The action output stream is not fed back into the model as it is redundant with the text stream input.

\looseness=-1
\textbf{Lookahead second text stream.}
The action stream allows the model to predict the next word position,
although the model has no knowledge of its content for making that decision. The delay between text and audio only provides context for the audio generation, however, the decision on where to insert pauses and words has no such context. Given a sequence of words $m_1, m_2, \ldots$, the lookahead text stream feeds the tokens of the words $m_{i + l}$ to the backbone while the primary text feed contains the tokens of words $m_{i}$.

\looseness=-1
\textbf{Speaker conditioning.} 
We provide speaker embeddings for up to 5 speakers.
Each speaker is represented by a 10s audio extract of the same speaker outside of the training segment. Speakers are identified using the diarization tool Pyannote~\citep{Bredin23} in the training data. If more than 5 speakers are present in the segment, only 5 at random are kept for the speaker embeddings. If less than 5 speakers are present, the remaining speaker slots are filled with learnt padding values. Each speaker audio extract is encoded with a \emph{speaker encoder} and results in a speaker embedding with a fixed dimension. We concatenate the speaker embedding from the different speakers, sum them with an absolute positional embedding, and feed them through cross-attention layers to the backbone. The speaker encoder has the same architecture as the encoder of the Mimi codec, and is initialized with its weights. We keep the weights of the convolutional layers frozen for stability, but let its Transformer layers be fine-tuned in an end-to-end fashion with the language model conditioned on it. 

\textbf{Change of turn tokens.}
We indicate change of turns between the first speaker in the speaker embedding, called the \emph{main speaker}, and any other speaker. When the main speaker starts talking, their first word is prefixed with a special \spectoken{MAIN} token in the text stream. When another speaker starts speaking after the main speaker, a special \spectoken{OTHER} token is inserted.
At inference time, we can thus make controllable dialogs by feeding the model with speaker embeddings for the two speakers, and controlling the change of turn by inserting the \spectoken{MAIN} and \spectoken{OTHER} special tokens.

\textbf{Classifier free guidance.}
We use classifier free guidance (CFG) ~\citep{ho2022classifier,audiogen}, both with respect to the speaker conditioning, and also with respect to the text, that is, we replace at inference time the logits for a given timestep $t$ and codebook index $q$, given $\alpha \geq 1$, with
\begin{equation}
\label{eq:cfg}
    l_{t, q} = l^{\emptyset}_{t, q} + \alpha \left(
    l^{\text{text,speaker}}_{t, q} - 
    l^{\emptyset}_{t, q}
    \right),
\end{equation}
where $l^{\emptyset}_{t, q}$ are the logit estimates obtained by feeding no text, action or lookahead inputs to the model, and no speaker embedding, and 
$l^{\text{text,speaker}}_{t, q}$ are the conditioned logits estimates.
No CFG is applied on the action stream logits. The model is trained with an independent dropout of 20\% on the speaker embedding and on the input text. Unless stated otherwise, we use $\alpha = 2$.

An overview of the whole DSM-TTS inference process is shown in Figure~\ref{fig:tts_inference}.

%% file: 80_appendix.tex
\appendix

\section{ASR Training Data}
\label{sec:app:data}
\subsection{Short-form finetuning data}
\label{sec:app:short-form}
This collection includes: LibriSpeech~\citep{librispeech}, VoxPopuli~\citep{voxpopuli}, GigaSpeech~\citep{gigaspeech}, AMI~\cite{ami-carletta2007,ami-renals2007}, SPGISpeech~\citep{spgispeech}, TED-LIUM~\citep{tedlium}, and Earnings22~\citep{earnings22}.
We filter out examples where Whisper had a high character error rate. %
With LibriSpeech (LS), we use timestamps provided by Montreal Forced Aligner~\citep{montreal_forced_aligner}, as prepared by~\cite{lugosch2019}. For the rest of the datasets, we use timestamps provided by Whisper (see Section~\ref{sec:dsm_asr}). Importantly, at this stage, examples are short: only 0.5\% of examples are longer than 30.5s. This data mixture totals 28k hours.

Table~\ref{tab:license_asr} summarizes licenses for the datasets used at this stage.

\begin{table}[h]
    \centering
\begin{tabular}{ll}
\toprule
\textbf{Dataset} & \textbf{License} \\
\midrule
AMI & CC-BY-4.0 \\
EARNINGS22 & CC-ShareAlike-4.0 \\
GigaSpeech & Apache 2.0 \\
LibriSpeech & CC-BY-4.0 \\
SPGISpeech & Non-commercial use \\
TED-LIUM & BY-NC-ND 3.0 \\
VoxPopuli & Creative Commons Zero \\
\bottomrule
\end{tabular}
    \caption{Licenses for the public datasets we used for \oursASR{} finetuning.}
    \label{tab:license_asr}
\end{table}
\subsection{Long-form finetuning data}
\label{sec:app:long-form}
First, we concatenate utterances in LibriSpeech~\citep{librispeech} to form segments of up to 6 minutes long ($\approx$ 1k hours). %
Second, we use a collection of synthesized dialogs, each dialog lasting for 5 minutes ($\approx 22$k hours). Those examples are produced by a preliminary version of \oursTTS. %
We combine these datasets with the short-form finetuning dataset in a weighted mixture, with respective weights 8:1:1 (i.e., a sample from LibriSpeech is 8x more likely to be included in a batch than from the short-form data mixture).

\section{Size ablations for ASR}
In the main text, we evaluated \oursASR{} model with 2.6B parameters. This model size might not be practical for all applications, so we additionally trained a smaller model with 300M parameters, using the same protocol, data, and vocabulary.

In Table~\ref{tab:300m-ablation}, we  compare its performance to several reference models. From this comparison, we see that \oursASR{} with 300M parameters  (350M including Mimi encoder) achieves considerably better performance than streaming and non-streaming versions of \textit{Whisper Medium}, which have 760M parameters. Unsurprisingly, the smaller version of \oursASR{} performs worse than a 2.6B version.

Finally, we note that these experiments suggest that the approach of \oursASR{} works well for models of smaller sizes, too.

\section{ASR: Efficiency as a function of batch size}
In Table~\ref{tab:bsz-throughput}, we report how the efficiency metrics (RTF and throughput) change in function of the batch size used. In these measurements, we use the same \oursASR{} as in the main text.

\begin{table}[t]
\centering
\caption{{\oursASR{}: Real-time Factor and Throughput in function of batch size}.}
\label{tab:bsz-throughput}
\begin{sc}
\begin{tabular}{ccc}
\toprule
\textbf{Batch size} & \textbf{RTF} $\uparrow$ & \textbf{Throughput} $\uparrow$ \\
\midrule
1 & 6.9 & 6.9 \\
32 & 4.4 & 141.4 \\
64 & 3.5 & 224.0 \\
256 & 1.49 & 380.1 \\
\bottomrule
\end{tabular}
\end{sc}
\end{table}

\begin{table}[t]
\centering
\caption{\textbf{Short-form ASR performance across model sizes}. We report Word Error Rates (WER, \%) for a \oursASR{} model with 300M parameters, along with reference models.}
\label{tab:300m-ablation}
\begin{sc}
\resizebox{\textwidth}{!}{
\begin{tabular}{lc|cccccccc}
\toprule
\textbf{Model} & \textbf{Avg.} %
& \textbf{AMI} & \textbf{Earnings22} & \textbf{Gigaspeech} & \textbf{LS Clean} & \textbf{LS Other} & \textbf{SPGISpeech} & \textbf{TED-LIUM} & \textbf{Voxpopuli}
\\
\midrule
\multicolumn{10}{c}{\textit{Non-streaming}} \\
\midrule[0.3pt]
Whisper medium.en & 8.1 %
& 16.7 & 12.6 & 11.0 & 3.0 & 5.9 & 3.3 & 4.1 & 9.6 \\
\midrule
\multicolumn{10}{c}{\textit{Streaming}} \\
\midrule[0.3pt]
Whisper medium.en & 9.0  & 22.1 & 13.4 & 10.4 & 3.0 & 6.2 & 3.7 & 4.7 & 8.6 \\
\midrule[0.01pt]
\oursASR{} 300M & {8.2} & {16.4} & {13.9} & {11.1} & {2.2} & {6.8} & {2.7} & {4.2} & {8.4} \\

\oursASR{} 2.6B & {6.4} & {12.2} & {11.0} & {9.8} & {1.7} & {4.3} & {2.0} & {3.4} & {6.8} \\
\bottomrule
\end{tabular}
}
\end{sc}
\end{table}

\section{Evaluation TTS Data}
\label{sec:app:eval_data}
For evaluating our data, we use both monologue and dialog scripts. 
We do not have reference audio to compare to, although we can still compare models to one another. The datasets and evaluation code are available at \href{https://github.com/kyutai-labs/tts_longeval}{github.com/kyutai-labs/tts\_longeval}.

\paragraph{Monologues.}
Those are taken from news articles of the NTREX-128 dataset~\citep{federmann-etal-2022-ntrex}, giving us 123 articles per language. Note that those articles are already split in sentences, which we leverage when evaluating models that do not support long form generation.

\paragraph{Dialogs.}
Dialogs are generated using the \texttt{mistral-large-latest} model through their API\footnote{https://docs.mistral.ai/api/}. The scripts fall in 3 categories: daily life, technical, and number-heavy. For daily life scripts, the LLM API is first asked to come up with a number of situations that could arise in daily interactions. For each one, it is then tasked with generating a script in a given language and with a target number of turns. We try to generate 50 daily dialogs per language (due to failures in the generation, only 97 in total are available). For technical topics, we follow a similar approach, except the LLM API is given a specific topic (technical, but also person, piece of art, movie,  etc.) to discuss instead. We again have 50 technical dialogs per language. Finally, for number-heavy dialogs, we use 12 hand written prompts, such as ``\emph{PERSON\_A and PERSON\_B discuss the chronology of various events during the middle age}'', which we use for generating scripts in both English and French. In total we thus have nearly 112 dialogs per language. Examples are provided in Figures~\ref{fig:dialog-example-daily} and \ref{fig:dialog-example-tech}.

\paragraph{Voices.}
For voice conditioning, we use samples from the test set of VCTK~\citep{Yamagishi2019CSTRVC} (ODC-BY license) for English, keeping only one utterance per speaker whose duration is longer than 7 seconds, and less than 10 seconds. We kept 50 voices.
For French, we combine speakers from the test and valid sets of CML~\citep{Cmltts2023} (CC-BY 4.0), applying the same criterion. This gives us 35 voices.
Pairs of voices are randomly assigned to each script, although the mapping is fixed across evaluation runs.

\section{Timestamp predictions}
\label{sec:app:timestamps}

We adopt evaluation metrics from~\cite{crisper}. Firstly, we compute an F1 score by defining true positives as words that match a reference word and overlap with it temporarily; false positive words are predicted but either do not match or do not overlap with a reference word; and false negatives are the reference words that do not have a match in content or have no temporal overlap with predicted words. The temporal overlap happens if the predicted timestamps for the word start and word end fall within a collar distance from the true events. Since this F1 metric considers overlap in a binary way, \cite{crisper} complement it with a more nuanced measure, mean Intersection over Union (mIoU), which additionally accounts for the relative duration of the temporal overlap. %
As the evaluation dataset, use the same time-aligned LibriSpeech test-clean as for latency measurement. 
Results are shown in Table~\ref{tab:alignments}.

\section{\oursASR{}: speech representation}
One natural question is whether \oursASR{} is hindered by using discretized speech representation. In order to answer this question, we use the fact that the models are trained using quantizer dropout: at training time, after a randomly selected cut-off level, all quantization levels are zeroed out~\citep{soundstream,moshi}. We exploit this by running a series of evaluations of the same model on LibriSpeech test-clean~\citep{librispeech}, while systematically dropping quantizers at different cut-off levels.
We observe that having fewer than 24 quantizers comes at a cost on WER; however, having more than that is not beneficial. Thus we conclude that having even more nuanced representation (e.g., continuous embeddings), might not lead to extra gains.

\begin{table}[t]
\centering
\caption{\textbf{Precision of predicted timestamps}. F1 and mIoU metrics, calculated on LS test-clean.}
\label{tab:alignments}
\begin{sc}
\begin{tabular}{lcccccc}
\toprule
Metric & {CrisperWhisper} & {Whisper large-v3} & \oursASR \\
\midrule
F1 & 0.85 & 0.22 & 0.73 \\
mIoU & 0.65 & 0.30 & 0.54\\
\bottomrule
\end{tabular}
\end{sc}
\end{table}

\section{TTS Baseline evaluations}
\label{sec:app:tts_baselines}
We present how the baselines presented in Section~\ref{sec:experiments:tts} were evaluated.
All baselines are given the same audio conditioning extracts of no more than 10 sec. for each speaker. The code for running the baselines is available at \href{https://github.com/kyutai-labs/tts_longeval}{github.com/kyutai-labs/tts\_longeval}.

\textbf{ElevenLabs.} For monologues, we directly feed the entire script. For dialogs, we feed turns one by one, with no context (except for the speaker conditioning).

\textbf{Dia.} Dia was only trained to generate short audio extracts. For dialogs, we provide the turns used for speaker conditioning, and the last turn that was generated (e.g. from the other speaker than we are currently generating). We take care to provide them in such an order that speaker alternate. Even with limited context, generation sometimes failed, in which case we revert to generating the segment with no context
(always keeping the speaker conditioning).

\textbf{Orpheus.} For monologues, we generate the sentences one by one
with a context given by the speaker conditioning audio sample, along with 
one sentence of context. When that fails, we only give the audio conditioning sample. For dialogs, as Orpheus is single speaker, we generate the turns one by one, only with the speaker conditioning as context.

\textbf{CSM.} CSM normally only supports dialogs. For monologues,
we have to repeat twice the same speaker in a row, which is out of distribution. We experimented with various parameters (deduplicating
the speaker conditioning to pretend they are two speakers, different context size) and found the best results in terms of WER was obtained
by keeping a single speaker conditioning, providing no extra past context, and still having two segments in a row with the same speaker id.
For dialogs, we start with giving the turns used for speaker conditioning. Then we experimented with giving no context (short form), or giving up to 6 turns of context (long form). In case where the context would be too long and CSM raised an error, we would try again with a shorter context.

\section{Human evaluations}
\label{sec:app:human_evaluations}

\subsection{Models and dataset}
The models are evaluated using the same parameters as described in Section~\ref{sec:app:tts_baselines}. For CSM, we only evaluate the version with no context which was giving the best WER.
We use a subset of the data provided in Section~\ref{sec:app:eval_data},
namely 15 monologues, and 9 dialogs (3 from each category) per language. For each script, we rate both the first 30 seconds, and the last 30 seconds separately.

\subsection{Subjective evaluation tasks}

Raters were payed on average 9.3£ per hour.

\textbf{Quality.}
We also organize a MUSHRA style study for evaluating the audio quality, although with no explicit reference. For each script, the rater is presented with all the models' generations, and must note each one on a scale from 1 to 100. In particular, the instruction is as follows:
``\emph{How would you rate the overall quality of these audio clips?
Consider aspects such as clarity, balance, richness, and naturalness}''.
We report the aggregated ratings, with the confidence intervals being the plus and minus the standard deviation of the mean estimator.
For each language, 800 scripts were rated (each time covering all methods).

\textbf{Speaker similarity.}
We present the rater with the audio used for speaker conditioning (either single speaker or the concatenation of both speakers for dialogs), along with two generations from two random models given the same script. The rater must choose which extract has speakers sound the most like the reference audio. The instruction is as follow:
``\emph{The reference contains the voices of one or two speakers. The audio files below it contain either a monologue or a dialog. Which voices sound more like the speakers in the reference?}''
The win-rate is summarized as an \emph{Elo score}, as described in the next paragraph. 1600 pairs were rated per language.

\textbf{Bayesian Elo Score.}
The Elo score allows the ranking of models based on some pairwise comparisons of audio samples. Given two models $A$ and $B$, the probability that $A$ is preferred over $B$ is:
\begin{equation}
    P(A>B) = \frac{1}{1 +10^{(E_B-E_A)/400}},
    \label{eq:elo}
\end{equation}
where $E_A$ and $E_B$ are the Elo scores of each model. 
Unlike a traditional Elo score, the Bayesian Elo score uses a Gamma prior,
so that one can derive confidence intervals over the posterior distribution.

We denote $S_A = 10^{(E_A - c)/400}$, with $c$ freely chosen in $\reals$. Then eq. \eqref{eq:elo} becomes
\begin{equation}
    P(A>B) = \frac{S_A}{S_A+S_B},
\end{equation}
which is a Bradley-Terry \citep{bradley_terry} model. We use the iterative by \citet{carondoucet} where $S_A^0$ follows a Gamma prior with parameters $\alpha^0, \beta^0$. 
By denoting $w_A$ the number of times where method $A$ won against any other method and $w_{AB}$ the number of times where $A$ and $B$ are compared, $S_A^t$ is computed with the following update rule:
\begin{equation}
    S_A^{t+1} = \frac{\alpha^0 +w_A}{\beta^0 +\sum_{B \neq A}\frac{w_{AB}}{S_A^t+S_B^t}},
\end{equation}
given as the mean of the Gamma distribution with updated parameters $\alpha_A^{t + 1}$, $\beta_A^{t +1 }$ given by
\begin{equation}
    \alpha_A^{t+1} = \alpha^0 + w_A,\qquad
    \text{and}\qquad \beta_A^{t+1} = 
    \beta^0 + \sum_{B\neq A} \frac{w_{A,B}}{S_A^t + S_B^t}.
\end{equation}
Iterating over $t$ allows reaching a fix point, we run 30 of them,
once we have collected all the pairs.
We use $\alpha=0.1, \beta=0.1, c=2000$ so that, in absence of any data, $S_A=1$ and $E_A = 2000$. Confidence intervals are 95\% confidence interval according to the posterior.

\section{Latency, RTF, and throughput}
\label{sec:app:latency}

We report in Table~\ref{tab:latency} latency, RTF, and throughput
for various batch sizes for our model. Those numbers are obtained under
real workload, with no synchronization between the batch elements,
that is, requests for TTS arrive at the service at \emph{random times}.

\paragraph{Batch size of 1.}
The latency represents the time to first audio, including all computations. Given that we use a delay $\tau=2$ sec., or 25 steps, we need to account for the time that it takes to run those.
We have to account for two times: $d_b$ the duration it takes
to run a single step of the backbone, and $d_q$ the time it takes to run
the small Transformer over the $Q$ dimension. For the first 25 steps, we need only to run the backbone transformer, as no audio is produced.
When the batch size is 1, the time to first token is given by $(\tau f_r - 1) \cdot d_b + (1 + 2)\cdot (d_b + d_q)$.  The extra $2\cdot (d_b + d_q)$ comes from the fact that, following~\citet{moshi}, we use an \emph{acoustic delay} of 2 steps between the first codebook and the remaining ones.
With values of $d_b=5\,\text{ms}$, and $d_q=25\,\text{ms}$, we obtained the given results. The throughput and RTF are derived using only the time per step of $d_b + d_q$.
Note that even though the small Transformer over $Q$ is smaller in terms of dimensions and layers per codebook, it represents a significant amount of computation due to us using $Q=32$ codebooks, leading to a large number of kernels being scheduled.

\paragraph{Interleaved steps for higher batch sizes.} 
When needing to handle requests that are not in sync, with a larger batch size, we cannot just skip the small Transformer over $Q$, as not everyone will be at the same point in the generation. We use a 2-for-1 interleaving pattern, where if any of the batch items is in the initial phase,
we interleave two steps of backbone only, followed by a full step.
For items in the initial phase, this gives an effective time per step
of $(d_b + 2 \cdot d_q) / 3$.

\section{TTS results on LibriSpeech and Seed-en training on public datasets}
\label{app:sec:libri_seed}

\begin{table}
\centering
\caption{\textbf{Resilience of existing watermarking techniques.} 
We compare the resilience of open-source watermarking techniques when applied to synthesized audio.
After having applied the watermark, we encode and decode the audio using the Mimi codec~\citep{moshi} with 32 RVQ levels.
The Perth watermark was released at \href{https://github.com/resemble-ai/perth}{github.com/resemble-ai/perth}.}
\label{tab:watermark}
\begin{tabular}{l|rr}
\toprule
\textbf{Watermarking model} & \textbf{Detection rate} (\%) & \textbf{Detection rate after Mimi} (\%)\\
\midrule
AudioSeal~\citep{sanroman2024proactive} & 100.0\% & 45.2\%\\
Resemble AI Perth & 100.0\% & 0.0\%\\
Silent Cipher~\citep{singh24_interspeech} & 82.3\% & 0.0\%\\
\bottomrule
\end{tabular}
\end{table}

We experiment with training \oursTTS{} using only the publicly available datasets. 

\paragraph{Architecture and training.} This variant of the model uses a 300m backbone (24 layers with dimension 1024), $Q=16$ codebooks. The small Transformer over $Q$ is made more compact following~\citep{hibiki}, e.g. sharing weights for all codebooks from $[9-16]$, $[17, 24]$, along with using only 4 layers per  step. We use a delay of $\tau=1.28\,\text{sec.}$, or 16 steps.
It receives no speaker conditioning, instead it has a probability of 20\% of seeing the start of the audio with no text, and a probability of 10\% of having this prefix completely empty, which will be used for applying classifier-free-guidance~\citep{ho2022classifier,audiogen}. The model receives no text 20\% of the time.
The model is trained for 500k updates with a batch size of 64 on 16 H100 GPUs, using the same optimization parameters as described in Section~\ref{sec:hyperparams}.

\paragraph{Dataset} We use a similar mixture as described in Section~\ref{sec:app:short-form}, using the following datasets: AMI, EARNINGS22, GIGASpeech, SPGISpeech, TED-LIUM, VoxPopuli. Besides, we use LibriHeavy~\citep{kang2023libriheavy} and Emilia~\citep{he2024emilia}.
For LibriHeavy, we use the original formatted text. We filter out data with a high word error rate compared with Whisper transcripts. This totals to 88k hours of speech.

\paragraph{Evaluations}

We evaluate our approach on LibriSpeech test clean, with punctuation, following the same protocol as \citet{chen2024f5} (F5-TTS). When evaluating our model, we provide the speaker conditioning sample as a prefix.
We use CFG factor $\alpha = 3$.
We use the exact same evaluation set as F5-TTS, along with the same text normalization, ASR model (\texttt{whisper-large-v3}), and reuse their code for evaluating the speaker similarity to the original conditioning audio.
We similarly evaluate on the Seed test en dataset~\citep{anastassiou2024seed}, following~\citep{du2025cosyvoice}.
We provide support for those benchmarks in our evaluation codebase \footnote{\href{https://github.com/kyutai-labs/tts_longeval}{github.com/kyutai-labs/tts\_longeval}}.
We release the $Q=16$ version publicly\footnote{\href{https://huggingface.co/kyutai/tts-0.75b-en-public}{huggingface.co/kyutai/tts-0.75b-en-public}}. This model allows for voice cloning through prefixing, although when generating turn-by-turn in the same setup as Table~\ref{tab:longform_similarity}, with a score of 74.9\%, against 80.9\% for our main model. This level of speaker similarity is in line with some of the baselines like CSM and it thus seems safe to open source it.

\paragraph{Results}

Results are provided in Table~\ref{tab:tts-libri}.
We achieve a large improvement of 0.3\% of WER over the best F5-TTS model, however, with a small decrease in speaker similarity. While the diffusion based approach F5-TTS can generate faster a single script, it suffers from a higher latency, due to its non causal nature requiring the entire audio to be generated at once. Finally, we again show that the ease of batching of our methods allows for more than 100x throughput (duration of audio generated per second of computation) even for requests arriving unsynchronized, as explained in Section~\ref{sec:app:latency}.

Finally, we also provide results on the Seed test en dataset in Table~\ref{tab:tts-seed}. Our 16 codebooks model beats F5-TTS and is quite close to Cosyvoice 3-1.5B (RL)~\citep{du2025cosyvoice}, while being twice as small, and not requiring a reinforcement learning based fine tuning stage.

\section{On the efficacy of watermarking for TTS}
\label{app:sec:watermark}

We challenge the practice of relying on watermarking to limit potentially negative usage of TTS models. First, when open sourcing such a model, the watermarking stage can be disabled in the code. Besides, even if the watermark was built in the model, we noticed that most existing state of the art watermarking methods would almost completely disappear after a single round of encoding/decoding with Mimi with 32 codebooks, e.g. with minimal distortions. Results are provided in Table~\ref{tab:watermark}. Only AudioSeal~\citep{sanroman2024proactive} is still detected half of the time. We believe this is because Mimi was not released at the time those watermark models were trained, and in particular, Mimi was trained with no spectrogram or waveform reconstruction loss, giving more freedom to how the input can be resynthesized, which the watermarking technique did not account for. This shows how fragile those approaches can be in the face of future innovation. The search for a reliable watermarking method remains an open and important question.

\section{Supplementary ablations on DSM-TTS}
\label{app:sec:ablations}
We now provide supplementary ablations.

First we compare training \oursTTS{} with Mimi and with EnCodec~\citep{encodec} in Table~\ref{tab:tts-encodec}. Training and evaluation are performed as in Section~\ref{app:sec:libri_seed}. We notice that while worse than with Mimi, our method generalizes well to other codecs. 

Second, we study the impact of the lookahead and action stream presented in Section~\ref{sec:dsm_tts} in Table~\ref{tab:tts-variants}. We train and evaluate as in Section~\ref{sec:experiments:tts}, either without a lookahead stream, or simply reusing the original \oursTTS{} model, we ignore its action prediction and instead use a fixed padding strategy between words: we either force a fixed amount of padding after the start of the word, or a fixed amount of padding after its last text token. We notice that not using the lookahead has limited impact on the speaker similarity, but deteriorates the WER. On the other hand, using a fixed padding pattern has a clear impact on the speaker similarity, likely due to the fixed and unnatural prosody.

\begin{table}[t]
\caption{Comparing \oursTTS{} trained with Mimi and with EnCodec~\citep{encodec}. The number of RVQ levels in EnCodec was selected to roughly match the number of tokens per second with Mimi, keeping still a sufficient number to ensure decent quality. Training and evaluations are done as in Table~\ref{tab:tts-libri}.}
\label{tab:tts-encodec}
\resizebox{\textwidth}{!}{
\begin{tabular}{l|llll|ll}
\toprule
\textbf{Model}                & \textbf{RVQ levels} & \textbf{Frame rate} & \textbf{Tokens per sec.} & \textbf{Bandwidth} & \textbf{WER} & \textbf{Speaker Sim.} \\
\midrule
F5-TTS               & -          & -          & -               & -         & 2.42 \%           & 0.66                       \\
DSM-TTS with Mimi    & 32         & 12.5 Hz    & 400             & 4.4 kbps  & 1.68 \%           & 0.71                       \\
DSM-TTS with Encodec & 8          & 75 Hz      & 600             & 6 kbps    & 2.45 \%           & 0.68                      \\
\bottomrule
\end{tabular}}
\end{table}

\begin{table}[t]
\caption{We compare variants of \oursTTS{}, in particular without a lookahead stream, and with a fixed padding strategy between words, e.g. without an action stream. For the fixed padding, we either force a fixed amount of padding after the start of the word, or a fixed amount of padding after its last text token. Training and evaluation is done as in Section~\ref{sec:experiments:tts}.}
\label{tab:tts-variants}
\begin{tabular}{lllll}
\toprule
\textbf{Model variant}                       & \textbf{WER English} & \textbf{WER French} & \textbf{Spk. Sim. English} & \textbf{Spk. Sim. French} \\
\midrule
DSM-TTS baseline            & 1.60\%      & 3.02\%     & 0.743             & 0.745            \\
No lookahead                & 3.51\%      & 3.25\%     & 0.743             & 0.746            \\
Fixed padding, 4 from start & 2.86\%      & 3.60\%     & 0.694             & 0.690            \\
Fixed padding, 5 from start & 2.69\%      & 3.48\%     & 0.691             & 0.700            \\
Fixed padding, 2 from end   & 2.32\%      & 3.13\%     & 0.715             & 0.698           \\
\bottomrule
\end{tabular}
\end{table}

\begin{figure}
    \centering
    \begin{tcolorbox}[width=\textwidth, colback=white, colframe=black, boxrule=0.8pt, fonttitle=\bfseries, title={Dialog example, daily life}]
[LIU]: Hey Jean, how's it going with the studies?

[JEAN]: Not bad, Liu. Just trying to keep up with everything.

[LIU]: I hear you. Have you tried any study apps lately?

[JEAN]: Not really. I've been sticking to the old pen and paper method.

[LIU]: Oh, you should try this new app. It's really helpful.

[JEAN]: What's it called?

[LIU]: It's called "StudyBuddy." It helps you organize your notes and has flashcards too.

[JEAN]: Sounds interesting. How does it work?

[LIU]: You can upload your notes, create flashcards, and it even has a quiz feature.

[JEAN]: That sounds really useful. I'll give it a try.

[LIU]: Great! I think you'll find it really helpful.

[JEAN]: Alright, I'll download it tonight.

[LIU]: Awesome. Let me know how it goes.

[JEAN]: Will do. Thanks for the recommendation, Liu.

[LIU]: No problem, anytime.

[JEAN]: Hey, I just downloaded the app. It looks pretty good.

[LIU]: Glad to hear it. Have you tried any of the features yet?

[JEAN]: I uploaded some notes and made a few flashcards. It's really user-friendly.

[LIU]: That's great to hear. I find the quiz feature really helpful too.

[JEAN]: Yeah, I'm going to try that next. Thanks again, Liu.

[LIU]: You're welcome, Jean. Good luck with your studies!

[JEAN]: Thanks! Talk to you later.

[LIU]: Take care, Jean. See you in class!
    \end{tcolorbox}
    \caption{Example of dialog generated with an LLM to serve as evaluation
    data. This dialog covers daily life topics.}
    \label{fig:dialog-example-daily}
\end{figure}

\begin{figure}
    \centering
    \begin{tcolorbox}[width=\textwidth, colback=white, colframe=black, boxrule=0.8pt, fonttitle=\bfseries, title={Dialog example, daily life}]
[LUIS]: Mrs. Al-Falasi, have you heard of hydroxyapatite before?

[MRS AL-FALASI]: Yes, Luis. I know it's related to bones and teeth, but that's about it.

[LUIS]: Great start! Hydroxyapatite is a naturally occurring mineral form of calcium apatite. It has the formula Ca5(PO4)3(OH).

[MRS AL-FALASI]: That sounds quite complex. What makes it so important?

[LUIS]: It's the main mineral component of bones and teeth. It provides rigidity and strength.

[MRS AL-FALASI]: That makes sense. Is it used in any medical applications?

[LUIS]: Yes, synthetic hydroxyapatite is used in medical and dental applications for bone grafts and implants.

[MRS AL-FALASI]: Really? How does it work in those contexts?

[LUIS]: It has biocompatible properties, which means it's suitable for use in body tissue repair.

[MRS AL-FALASI]: That's fascinating. Are there any other uses for it?

[LUIS]: Absolutely. Hydroxyapatite can be used in coatings on orthopedic implants to promote bone growth.

[MRS AL-FALASI]: I didn't know that. It seems very versatile.

[LUIS]: It is. It's also used in toothpaste and mouthwash to help remineralize enamel.

[MRS AL-FALASI]: That explains why it's so beneficial for dental health. Anything else I should know?

[LUIS]: The material is effective in drug delivery systems, especially for bone diseases.

[MRS AL-FALASI]: That's impressive. Are there any non-medical uses?

[LUIS]: Yes, hydroxyapatite is utilized in chromatography for protein purification.

[MRS AL-FALASI]: Interesting. How about its sources? Where does it come from?

[LUIS]: It can be derived from natural sources such as animal bones or synthesized chemically.

[MRS AL-FALASI]: That's good to know. Any other applications I might not be aware of?

[LUIS]: Its porous structure makes it useful in filtering and ion-exchange applications.

[MRS AL-FALASI]: That's a lot of information. Thanks for explaining, Luis.

[LUIS]: You're welcome, Mrs. Al-Falasi. Hydroxyapatite is truly a remarkable material.
    \end{tcolorbox}
    \caption{Example of dialog generated with an LLM to serve as evaluation
    data. This dialog covers technical topics.}
    \label{fig:dialog-example-tech}
\end{figure}